\documentclass[10pt,twocolumn,letterpaper]{article}

\usepackage{cvpr}             

\definecolor{cvprblue}{rgb}{0.21,0.49,0.74}
\usepackage[pagebackref,breaklinks,colorlinks,allcolors=cvprblue]{hyperref}
\usepackage{multirow}
\usepackage{graphicx}  
\usepackage{tabularx}  
\usepackage{array}      
\usepackage{lipsum}     
\usepackage{ragged2e} 
\usepackage{balance}
\usepackage{subcaption}
\usepackage[most]{tcolorbox}
\usepackage{lmodern}
\usepackage{caption}
\tcbuselibrary{listingsutf8}

\def\paperID{*****} 
\def\confName{CVPR}
\def\confYear{2025}

\title{Differential Attention for Multimodal Crisis Event Analysis}

\author{
\and
{}
\and
Nusrat Munia\\
University of Kentucky\\
Lexington, KY, USA\\
{\tt\small nusrat.munia@uky.edu}
\and
Junfeng Zhu\\
Kentucky Geological Survey\\
Lexington, KY, USA\\
{\tt\small junfeng.zhu@uky.edu}
\and
{}
\and
{}
\and
Olfa Nasraoui\\
University of Louisville\\
Louisville, KY, USA\\
{\tt\small olfa.nasraoui@louisville.edu}
\and
Abdullah-Al-Zubaer Imran\\
University of Kentucky\\
Lexington, KY, USA\\
{\tt\small aimran@uky.edu}
}

\begin{document}
\maketitle
\begin{abstract}
Social networks can be a valuable source of information during crisis events. In particular, users can post a stream of multimodal data that can be critical for real-time humanitarian response. However, effectively extracting meaningful information from this large and noisy data stream and effectively integrating heterogeneous data remains a formidable challenge. In this work, we explore vision language models (VLMs) and advanced fusion strategies to enhance the classification of crisis data in three different tasks. We incorporate LLaVA-generated text to improve text-image alignment. Additionally, we leverage Contrastive Language-Image Pretraining (CLIP)-based vision and text embeddings, which, without task-specific fine-tuning, outperform traditional models. To further refine multimodal fusion, we employ Guided Cross Attention (Guided CA) and combine it with the Differential Attention mechanism to enhance feature alignment by emphasizing critical information while filtering out irrelevant content. Our results show that while Differential Attention improves classification performance, Guided CA remains highly effective in aligning multimodal features. Extensive experiments on the CrisisMMD benchmark data set demonstrate that the combination of pretrained VLMs, enriched textual descriptions, and adaptive fusion strategies consistently outperforms state-of-the-art models in classification accuracy, contributing to more reliable and interpretable models for three different tasks that are crucial for disaster response. Our code is available at \url{https://github.com/Munia03/Multimodal_Crisis_Event}.
\end{abstract}    
\section{Introduction}
\label{sec:intro}

\begin{figure*}
  \centering
  \resizebox{\linewidth}{!}{
  \includegraphics[width=0.95\linewidth]{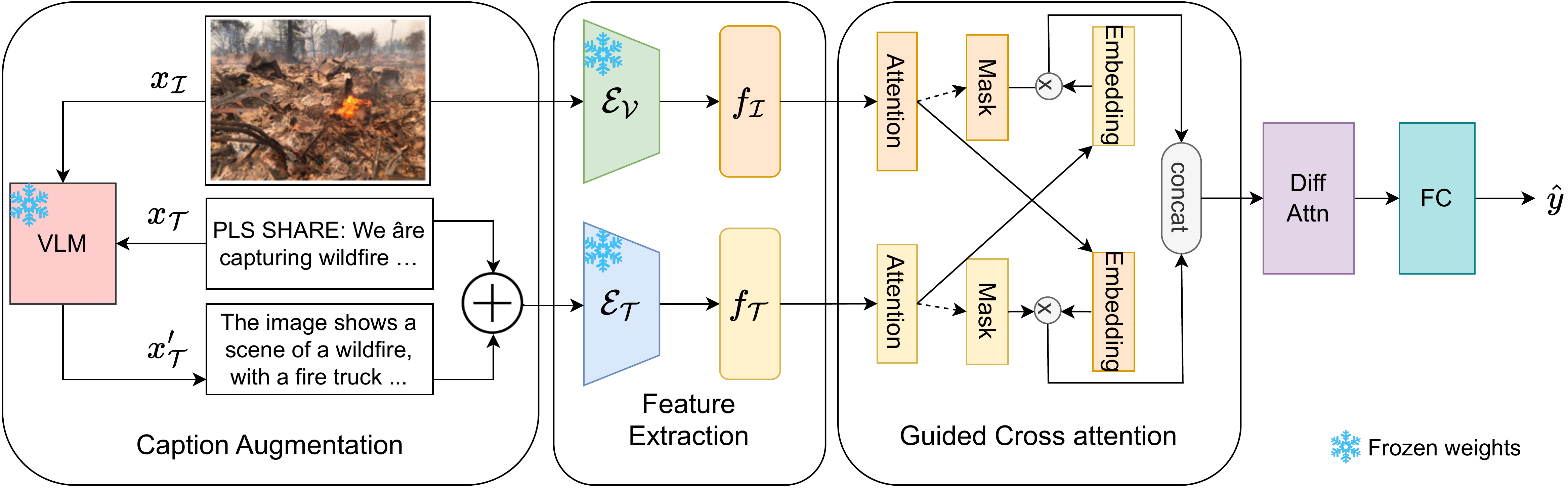}
  \hfill
  }
  \label{fig:archi}
  \caption{Proposed model architecture with VLM-infused Knowledge Fusion (KF) and Differential Attention with Guided Cross Attention.}
  \label{fig:archi}
\end{figure*}

During crisis events such as wildfires, hurricanes, floods, and tsunamis, people actively share updates, images, and videos on social networks. This creates a vast pool of data that can aid in humanitarian response and decision-making. Extracting important information from this ongoing stream of data can help make quick decisions and use resources more effectively. However, not all social media posts contain relevant or actionable information. It becomes essential to filter out non-informative content and identify meaningful posts that support crisis management efforts. 

Multimodal machine learning has shown immense potential in various applications, including sentiment analysis \cite{lai2023multimodal}, misinformation detection \cite{shu2022combating}, etc. Despite its advantages, multimodal learning remains challenging due to complex interactions between different modalities and the difficulty of aligning heterogeneous data sources \cite{baltruvsaitis2018multimodal, liang2022foundations}. Previous research has highlighted that training multimodal classification networks is often more difficult than their unimodal counterparts due to modality imbalance, misalignment, and varying levels of noise in modalities \cite{wang2020makes}. Addressing these challenges is crucial for building robust models that effectively support crisis response efforts.

Vision-language models (VLMs) have made significant strides in aligning image and text modalities across various tasks, including image captioning, visual question answering, and cross-modal retrieval. Learning to combine visual and text information improves integration and understanding, enhancing interpretability and context~\cite{radford2021learning, liu2023visual}. VLMs can create detailed and meaningful descriptions from images, which makes them especially useful for analyzing data related to text and images. We incorporate VLMs to generate detailed image descriptions, enriching textual information for multimodal learning.

Furthermore, to advance multimodal learning for the classification of crisis events, we combine the guided cross-attention with the differential attention mechanism \cite{ye2024differential} that refines the fusion process between textual and visual modalities. Unlike conventional attention mechanisms, which treat all modalities uniformly, our approach assigns adaptive attention weights based on the contextual importance of each modality in a given instance.
Our contribution is summarized below.

\begin{itemize}
    \item Adapting knowledge fusion by leveraging VLMs to enhance text-image alignment.
    \item Utilizing pre-trained models to facilitate enhanced feature representations without requiring further training.
    \item Exploring various fusion mechanisms to achieve better alignment of vision and text representations, thereby enhancing classification performance. 
    \item We validate our approach on the CrisisMMD benchmark dataset, where it consistently outperforms existing multimodal fusion techniques.
\end{itemize}

\section{Method}
\label{sec:method}

This section details the key components of our approach, including feature extraction and attention-based fusion. Fig.~\ref{fig:archi} depicts the proposed architecture that includes knowledge fusion, differential attention, and guided cross-attention. 

Given a crisis event data pair $ (x_I, x_T, y)$, where $x_I$ is the image, $x_T$ is the text, and $y$ is a task-specific class label, our objective is to improve classification performance by improving cross-modal interaction and knowledge fusion. 

First, we utilize a VLM to get image-specific text captions. We generate an instruction $I$ that includes the original text of the social media post $x_T$. We provide this instruction to the VLM (LLaVA) with the corresponding image to generate a descriptive caption of the input image $x_I$. The VLM model generates the image-specific caption ($x_T'$) based on instruction $I$. 
\begin{equation}
x_T' = \text{VLM}(I, x_I).
\end{equation} 

Next, we concatenate the newly generated caption with the original post's text to get the final text $x_T''$ that contains both the generated caption and image-specific information.
\begin{equation}
x_T'' = concat(x_T, x_T').
\end{equation} 
To obtain multimodal feature representations, we use CLIP's pretrained encoders (Fig.~\ref{fig:archi}). The image encoder extracts a feature representation $f_I$, and the text encoder generates an embedding $f_T$ as follows.
\begin{equation}
\label{eq:clip_feat}
f_I = \text{CLIP}{\text{image}}(x_I), \quad f_T = \text{CLIP}{\text{text}}(x_T'').
\end{equation}

We keep both encoders frozen during training to preserve the pretrained semantic alignment between image and text modalities. Then we apply the Guided Cross Attention module, similar to \cite{gupta2024crisiskan}. First, we apply self-attention separately to each modality. Given a feature vector $V$, self-attention is computed as follows.
\begin{equation}
\text{Self-Attn}(V) = \text{softmax} \left( \frac{V V^T}{\sqrt{d}} \right) V,
\end{equation}
\label{eq:self_attention}
where $d$ is the embedding dimension. This mechanism helps suppress noise and emphasize salient features within each modality. After applying self-attention on $f_I$ and $f_T$, we obtain new representations, given by
\begin{equation}
f_I = \text{Self-Attn}(f_I), \quad f_T = \text{Self-Attn}(f_T).
\end{equation}

Then, we calculate a new projection of $f_I$, $f_T$, and their attention mask as follows.
\begin{equation}
z_I = F(W_{I}^{T}f_I+b_I), \quad \alpha_I = \sigma(W_{I}^{'T}f_I+b_I),
\end{equation}
where $F(.)$ is the activation function and $\sigma$ is the Sigmoid function. Similarly, for text
\begin{equation}
z_T = F(W_{T}^{T}f_T+b_T), \quad \alpha_T = \sigma(W_{I}^{'T}f_T+b_I).
\end{equation}
Then, the attention mask of vision is multiplied by the projection of text, and the attention mask of text is multiplied by the projection of vision. In this way, both modalities have an enhanced feature representation with guided cross-attention. Next, we obtain the final representation from both modalities by concatenating the final vision and the text attention vectors $z = \text{concat}(\alpha_T.z_I, \alpha_I.z_T)$.

After getting the final representation by concatenation, we apply the differential attention mechanism to the final feature vector. For the differential attention mechanism \cite{ye2024differential}, given input $X \in \mathbb{R}^{N \times d_{\text{model}}}$, we get the projection matrices query, key, and value as follows.
\begin{equation}
[Q_1; Q_2] = X W^Q, [K_1; K_2] = X W^K, V = X W^V.
\end{equation}
Next, the differential attention is calculated using
\begin{equation}
\begin{aligned}
\text{DiffAttn}(X) & = \Big( \text{softmax} \Big(\frac{Q_1 K_1^T}{\sqrt{d}} \Big) \\
& - \lambda \, \text{softmax} \Big(\frac{Q_2 K_2^T}{\sqrt{d}} \Big) \Big) V,
\end{aligned}
\end{equation}

where $W^Q, W^K, W^V \in \mathbb{R}^{d_{\text{model}} \times 2d}$ are parameters, and $\lambda$ is a learnable scalar. Then we get the final refined representation $z'=\text{DiffAttn(z)}$. This final representation is passed lastly to a classification head for the prediction of crisis events: $\hat{y} = \text{softmax}(FC(z'))$.

\section{Experimental Evaluation}
\label{sec:results}

\begin{table}[t]
    \centering
    \caption{Dataset distributions across all three tasks.}
    \resizebox{0.9\linewidth}{!}{
    \begin{tabular}{@{}lcccc@{}}
    \toprule
        \textbf{Task} & \textbf{Train} & \textbf{Validation} & \textbf{Test} & \textbf{Total} \\
        \midrule
        Informativeness & 9599 & 1573 & 1534 & 12706 \\
        Humanitarian & 2874 & 477 & 451 & 3802 \\
        Damage Severity  & 2468 & 529 & 529 & 3526 \\
        \bottomrule
    \end{tabular}}
    \label{tab:dataset_distribution}
\end{table}

\begin{table*}[t]
    \centering
    \caption{Performance comparison of using pre-trained VLMs as vision and text encoders, two Knowledge Fusion (KF) methods, and various fusion strategies on all three tasks.}
    \vspace{-0.3cm}
    \resizebox{\textwidth}{!}{%
    \begin{tabular}{@{}llll ccc c ccc c ccc@{}}
        \toprule
        \multirow{2}{*}{\textbf{Vision}}  & \multirow{2}{*}{\textbf{Text}} & \multirow{2}{*}{\textbf{KF}} & \multirow{2}{*}{\textbf{Fusion}} & \multicolumn{3}{c}{\textbf{Task 1}} & \phantom{a} & \multicolumn{3}{c}{\textbf{Task 2}} & \phantom{a}  & \multicolumn{3}{c}{\textbf{Task 3}}\\
        \cmidrule{5-7} \cmidrule{9-11} \cmidrule{13-15} 
         &  &  & & \textbf{Accuracy} & \textbf{Macro F1} & \textbf{W-F1}  && \textbf{Accuracy} & \textbf{Macro F1} & \textbf{W-F1} && \textbf{Accuracy} & \textbf{Macro F1} & \textbf{W-F1} \\
        \midrule
        DenseNet & - & - & - 
        & 82.89 & 80.81 & 82.98 
        && 86.25 & 61.39 & 85.78 
        && 62.57 & 49.63 & 62.00 \\
        - & Electra & Wiki & - 
        & 84.64 & 81.70 & 84.22 
        && 87.36 & 60.46 & 87.45 
        && 62.45 & 50.37 & 62.63 \\
        DenseNet & Electra & Wiki & Guided CA 
        & 86.80 & 85.25 & 86.87 
        && 91.34 & 66.08 & 91.22 
        && 64.65 & 44.64 & 61.03
        \\
        \midrule
        DenseNet & Electra & Wiki & Cross Attention 
        & $87.32 $ & $85.71 $ & $87.36 $ 
        && $89.28 $ & $62.53 $ & $88.82 $ 
        && $63.07 $ & $42.97 $ & $59.32 $ \\
        DenseNet & Electra & Wiki & Guided CA+Self Attn 
        & $88.36 $ & $87.00 $ & $88.44 $ 
        && $91.43 $ & $60.25 $ & $90.75 $ 
        && $63.83 $ & $44.95 $ & $60.92 $\\
        DenseNet & Electra & Wiki & Cross Diff Attn 
        & $85.74 $ & $85.74 $ & $83.99 $ 
        && $86.55 $ & $51.42 $ & $85.51 $
        && $61.69 $ & $41.21 $ & $58.19 $ \\
        DenseNet & Electra & Wiki & Guided CA+Diff Attn 
        & $89.33 $ & $87.94 $ & $89.35 $ 
        && $91.58 $ & $57.44 $ & $90.68 $ 
        && $63.14 $ & $46.89 $ & $61.13 $ \\
        \midrule
        CLIP Vision & CLIP text & Wiki & Guided CA  
        & $90.57 $ & $89.05 $ & $90.45 $ 
        && $\textbf{94.02} $ & $\underline{70.95} $ & $\underline{93.65} $
        && $\underline{68.94} $ & $53.50 $ & $65.69 $ \\
        CLIP Vision & CLIP text & Wiki & Guided CA+Diff Attn 
        & $90.44 $ & $89.06 $ & $90.39 $ 
        && $93.72 $ & $\textbf{71.04} $ & $93.37 $
        && $68.68 $ & $53.04 $ & $65.55 $ \\
        CLIP Vision & CLIP text & LLaVA & Guided CA+Diff Attn 
        & $\underline{92.52} $ & $\underline{91.40} $ & $\underline{92.47} $ 
        && $93.87 $ & $68.87 $ & $93.44 $ 
        && $68.87 $ & $\underline{55.11} $ & $\textbf{66.62} $\\
        CLIP Vision & CLIP text & LLaVA & Guided CA  
        & $\textbf{92.91} $ & $\textbf{91.91} $ & $\textbf{92.89} $
        && $\underline{93.92} $ & $69.45 $ & $\textbf{93.69} $ 
        && $\textbf{69.00} $ & $\textbf{55.14} $ & $\underline{66.28} $ \\
        \bottomrule
    \end{tabular}
    \label{tab:task1_res}
    }
\end{table*}

\subsection{Data}
We evaluate our multimodal crisis classification model using the CrisisMMD dataset \cite{alam2018crisismmd}, a large-scale collection of Twitter posts shared during natural disasters in 2017. CrisisMMD includes posts from seven major natural disasters, including hurricanes, earthquakes, floods, and wildfires. The dataset is annotated for three classification tasks.  Task 1 aims to classify image-text pairs as either informative or noninformative. Task 2 focuses on five humanitarian classes: Infrastructure damage, Vehicle damage, Rescue efforts, Affected individuals (including injuries, deaths, missing persons, and found individuals), and Others. Task 3 is designed for severity assessment, classifying instances into Severe, Mild, and Little or no damage.
We follow a setup similar to the methodology in \cite{abavisani2020multimodal} for evaluation. Additionally, we consider only image-text pairs with identical labels to maintain consistency in classification tasks. The dataset distribution for all three tasks is shown in Table~\ref {tab:dataset_distribution} 
and sample data are shown in Fig.~\ref{fig:data_samples}.

\subsection{Experimental Setup \& Evaluation Metrics}
We trained our model on the three tasks using the following parameters: base learning rate $1 \times 10^{-3}$, batch size $32$, SGD optimizer, categorical cross-entropy loss function, and epochs $50$. We applied the teacher's training method, where we stopped training when the validation loss no longer decreased, with a patience of 5. We developed our models in Python using PyTorch and ran them on an Intel(R) Xeon(R) system with 128GB of RAM and two NVIDIA RTX A4000 GPUs. We evaluated our model's performance by calculating accuracy, macro F1-score, and weighted F1-score. We report the average scores after three runs of each experiment.

\subsection{Results and Discussion}
Table~\ref{tab:task1_res} reports the classification results for Task 1, Task 2, and Task 3. Initially, we experimented with unimodal models: image-only and text-only models. For all tasks, the combination of vision and text embeddings is found to play a critical role in enhancing the model's ability to classify and interpret multimodal data. Following CrisisKAN~\cite{gupta2024crisiskan}, we employed DenseNet \cite{huang2017densely} for vision embeddings and Electra \cite{clark2020electra} for text embeddings. For KF, we collected related information from Wikipedia. We experimented with various fusion strategies to integrate these text and image modalities. The fine-tuning of these models enables the model to learn task-specific features, but requires a significant amount of training data and computational resources. 
When using CLIP Vision and Text embeddings, we observed a significant improvement in performance across tasks. Unlike the fine-tuned DenseNet and Electra models, the CLIP-based models leverage frozen pre-trained representations. The strong performance of CLIP embeddings suggests that these pre-trained multimodal representations are highly effective at capturing semantic relationships between images and text, even without additional task-specific fine-tuning.
When comparing KF strategies, we evaluated two different sources of textual knowledge: Wikipedia-collected text and LLaVA-generated text. Our findings indicate that LLaVA-generated text consistently improves classification performance across the three tasks compared to Wikipedia-based knowledge. LLaVA-generated text is more contextually aligned with visual content, as LLaVA is trained to generate descriptions that are semantically relevant to the input image. This ensures that the additional textual information provided by LLaVA is highly relevant to the classification task, thereby improving the model’s ability to make accurate predictions.

Among the fusion strategies explored, the Guided Cross Attention (Guided CA) mechanism has already demonstrated superior performance in prior work \cite{gupta2024crisiskan}. By selectively focusing on the most informative aspects of both modalities, Guided CA improves the ability of the model to extract meaningful relationships between images and text, leading to improved classification accuracy. Our additional differential attention layer, along with Guided CA, aimed to enhance the model’s focus on salient features while reducing noise from less relevant information. Differential attention assigns greater importance to critical features that contribute significantly to classification predictions, thus improving interpretability and robustness. The enhanced attention mechanism likely helped the model capture subtle yet crucial relationships between visual and textual elements, leading to more precise predictions. However, for Tasks 2 and 3, the additional differential attention layer did not yield substantial improvements over the standard Guided CA approach. Only minor performance gains were observed. This suggests that for these tasks, Guided CA already effectively aligned multimodal features, and the differential attention layer did not introduce a significant advantage. One possible explanation is that Tasks 2 and 3 may involve less ambiguous decision boundaries, where the existing fusion strategy was already sufficient to capture the necessary cross-modal relationships.

\section{Conclusion}
\label{sec:conclusion}

We have explored multimodal learning approaches to improve the classification of crisis events by enhancing the alignment between vision and text representations. Our study demonstrated that leveraging large-scale VLMs, such as CLIP and LLaVA, enhances text-image alignment and improves classification accuracy. We showed that using pre-trained CLIP embeddings without fine-tuning outperforms conventional fine-tuned models. Additionally, our experiments with knowledge fusion strategies revealed that LLaVA-generated text provides richer contextual information than Wikipedia-derived knowledge, leading to better classification performance. We also explored various fusion mechanisms, demonstrating that guided cross-attention effectively integrates multimodal features. Although adding differential attention further refined the classification for some tasks, its impact varied depending on the characteristics of the dataset. In general, our findings emphasize the importance of multimodal fusion enhanced with VLM knowledge and robust attention mechanisms in crisis event classification, paving the way for more effective AI-driven humanitarian response systems. Our ongoing work focuses on the further refinement of fusion strategies and larger-scale validation.

\section{Acknowledgement}

This work is supported by a grant funded by the National Science Foundation (NSF) under Grant number 2344533.

{ \balance
    \small
    \bibliographystyle{ieeenat_fullname}
    \bibliography{main}
}

\clearpage                      
\appendix                      
\section{Appendix}
\begin{figure*}
  \centering
  \resizebox{\linewidth}{!}{
   \begin{tabular}{c c}
   \multirow{2}{*}{\subcaptionbox{\large Task 1: Informativeness}{\includegraphics[width=0.7\linewidth]{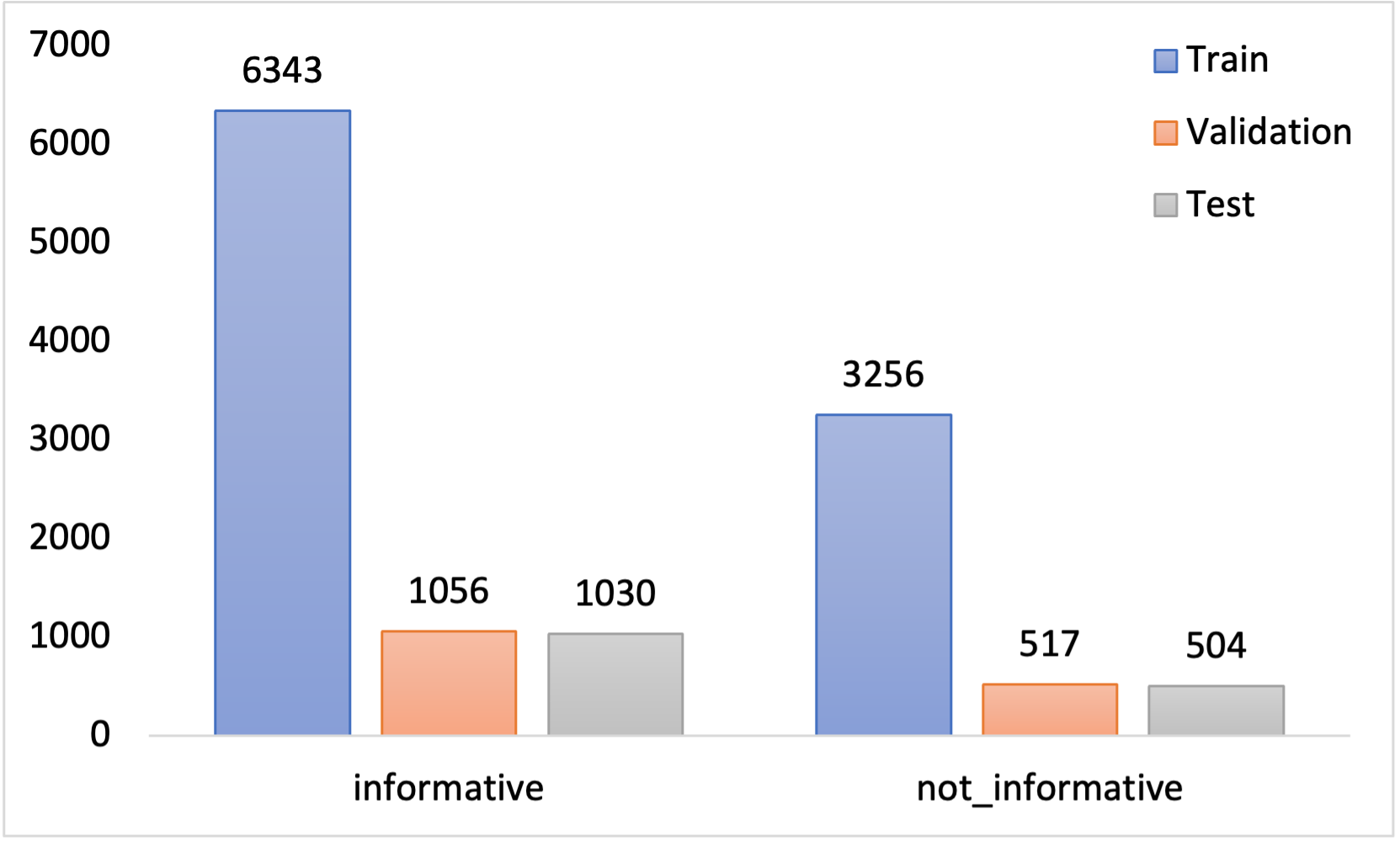}}}
    &
   \subcaptionbox{\large Task 2: Humanitarian}{\includegraphics[width=\linewidth]{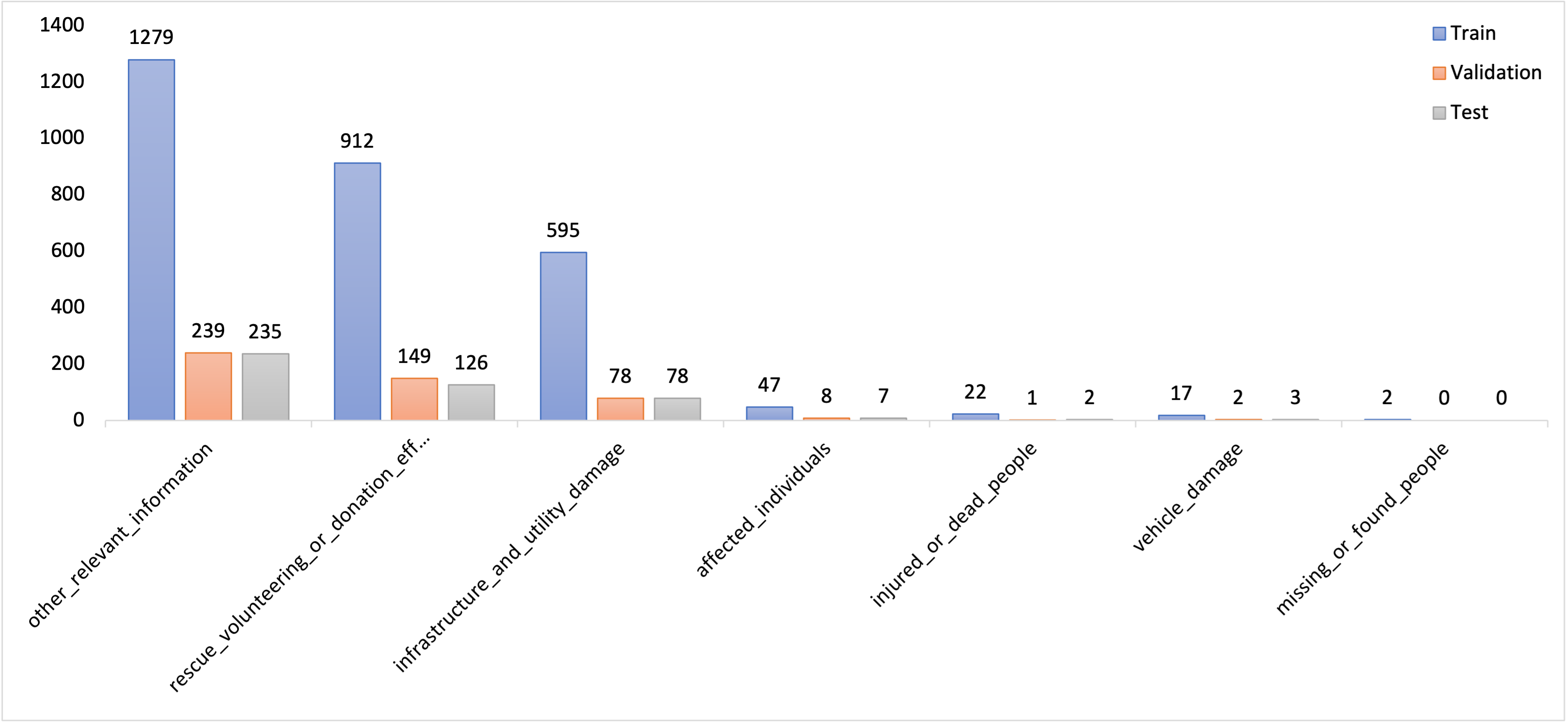}}
  \medskip\\
  &
  \subcaptionbox{\large Task 3: Damage severity}{\includegraphics[width=0.9\linewidth]{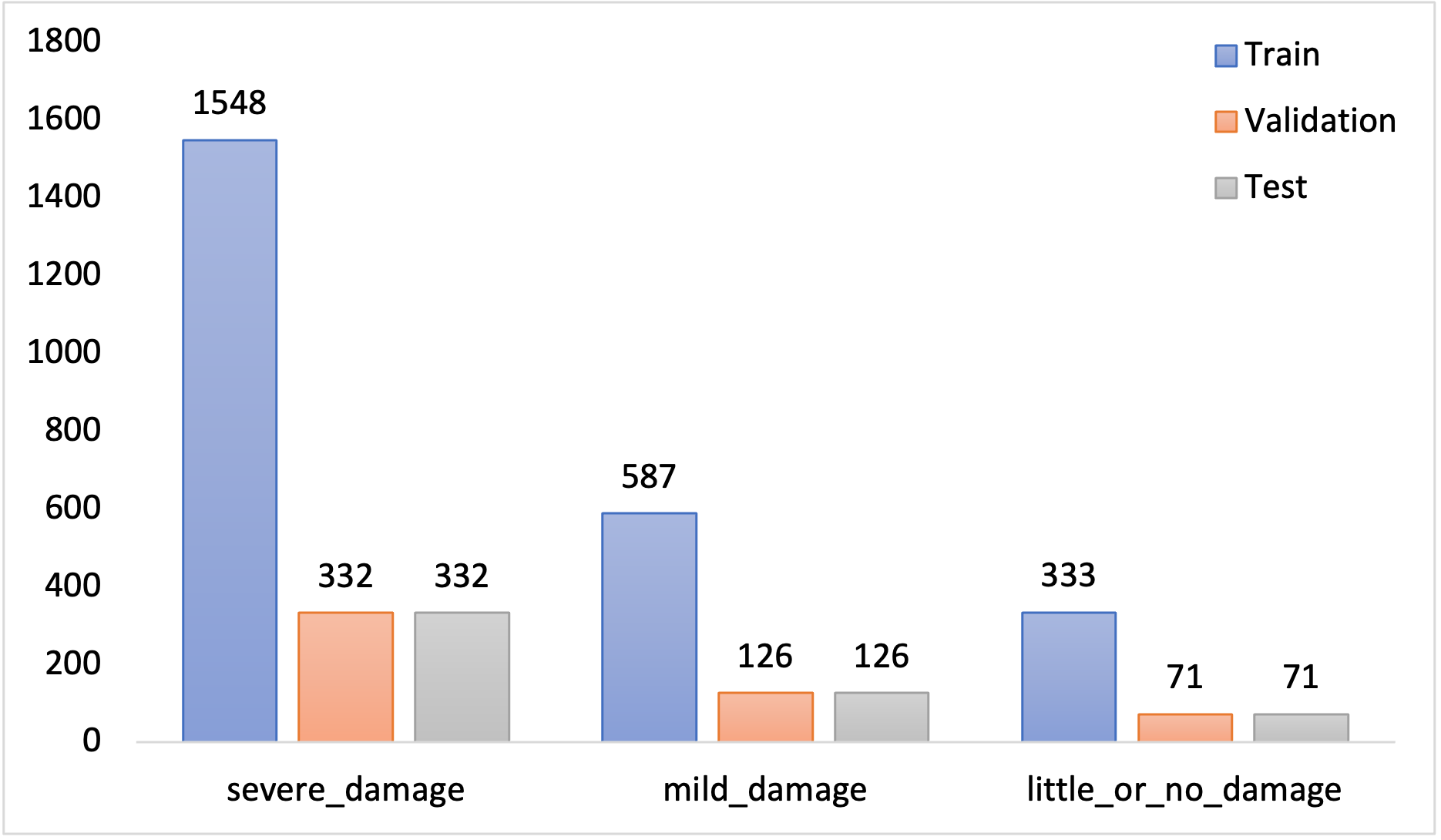}}
  \end{tabular}
  }
  \caption{Class-wise train/test/validation splits of the CrisisMMD dataset for all three tasks.}
  \label{fig:data_distribution}
\end{figure*}

In Fig.~\ref{fig:data_distribution}, we illustrate the number of samples for each class across the train, validation, and test sets for all three tasks. The figure clearly demonstrates a high degree of class imbalance in the CrisisMMD dataset. For Task 1, the distribution is skewed toward the non-informative class, which significantly outnumbers the informative samples. In Task 2, several classes contain very few instances. To mitigate this extreme sparsity, we grouped three underrepresented categories: affected individuals, injured or dead people, and missing or found people, into a single class during training. Similarly, Task 3 exhibits notable class imbalance, with the severe damage category having substantially more samples than the mild damage and little or no damage classes.

Fig.~\ref{fig:prompt} presents the full system prompt used to guide the LLaVA model in generating detailed and context-aware descriptions of disaster-related imagery. The prompt is structured using special tokens $(<|im\_start|>, <|im\_end|>)$ to delineate roles and message boundaries for system, user, and assistant interactions.

\begin{figure*}
\centering
\begin{tcolorbox}[
    colback=gray!10,
    colframe=black,
    colbacktitle=black,
    fontupper=\ttfamily\small,
    listing only,
    listing options={
        basicstyle=\ttfamily\small,
        breaklines=true,
        breakatwhitespace=false,
        columns=fullflexible,
        keepspaces=true,
        tabsize=2
    },
    enhanced,
    sharp corners=south,
    width=\textwidth,
    boxrule=0.8pt,
    top=1mm, bottom=1mm, left=2mm, right=2mm,
    boxsep=2pt
]
<|im\_start|>system\\
- You are LLaVA, a large multimodal assistant trained by UW Madison WAIV Lab.  \\
- You can understand and analyze visual content provided by the user and assist with a variety of tasks using natural language.  \\
- Follow the instructions carefully and provide detailed, context-aware explanations.  <|im\_end|>  \\
<|im\_start|>user \\ 
Given the caption: <\textit{tweet\_text}>, analyze the corresponding image and describe it in a very detailed and informative manner, focusing on crisis-relevant visual elements such as damage level, people, infrastructure, or rescue efforts. <\textit{image}> <|im\_end|>  
<|im\_start|>assistant
\end{tcolorbox}
\caption{Instruction prompt used for generating image description using the LLaVA model.}
\label{fig:prompt}
\end{figure*}

\begin{figure*}
    \centering
    \resizebox{\linewidth}{!}{
    \begin{tabular}{@{}m{0.1\linewidth} m{0.22\linewidth} m{0.15\linewidth} m{0.42\linewidth} m{0.11\linewidth} @{}}    
    \toprule
        \textbf{Task} & \textbf{Image} & \textbf{Tweet Text} & \textbf{LLaVA generated Text} & \textbf{Label} \\
        \midrule
        Task 1 & {\includegraphics[width=4cm,height=4cm]{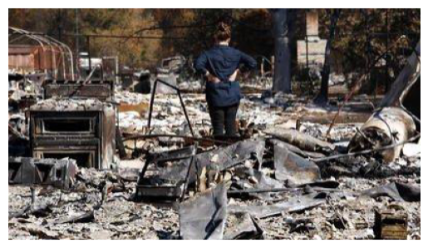}} 
        & Residents recall the night they had to flee the \#redwoodvalleyfire.  & The image shows a woman standing in a destroyed, burned-out building, surrounded by debris and rubble. The scene appears to be a result of a fire that has ravaged the area, leaving behind a desolate and destroyed landscape. The woman is likely a resident who had to flee the fire and has returned to the scene to assess the damage or to search for any belongings that might have been lost during the fire. The image captures the aftermath of a disaster, highlighting the destruction and the challenges faced by those affected by such events. 
        & informative \\
        \midrule
        Task 2 & {\includegraphics[width=4cm,height=4cm]{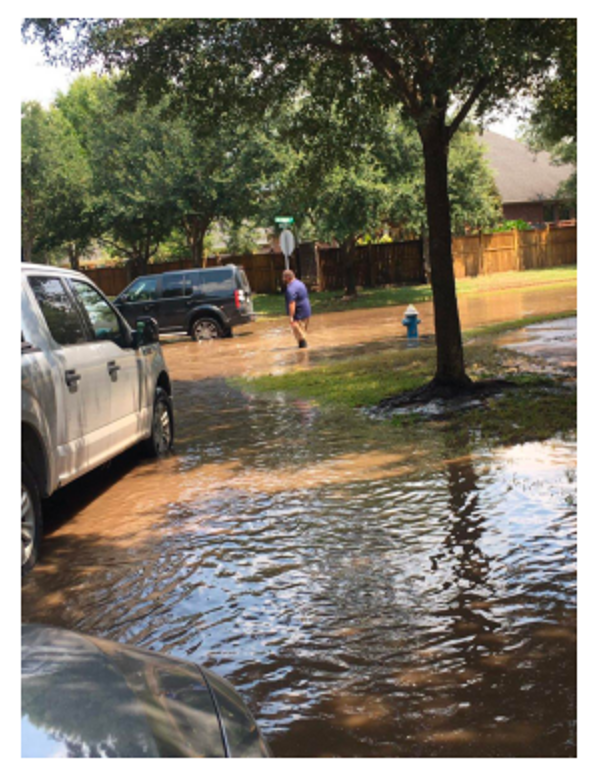}} & 
        Houston neighborhoods deep in \#HarveyRecovery \#txlege \#harvey 
        & 
        The image shows a flooded street in a Houston neighborhood, with cars and trucks parked on the side of the road. A man is walking through the flooded area, possibly navigating through the water to reach his vehicle. The scene is set in a residential neighborhood, and the flooding appears to be a result of the recent Harvey storm. The presence of a car and a truck in the image indicates that the residents are trying to cope with the aftermath of the storm and the ongoing recovery efforts.
        &
        infrastructure and utility damage \\
        \midrule
        Task 3  & {\includegraphics[width=4cm,height=4cm]{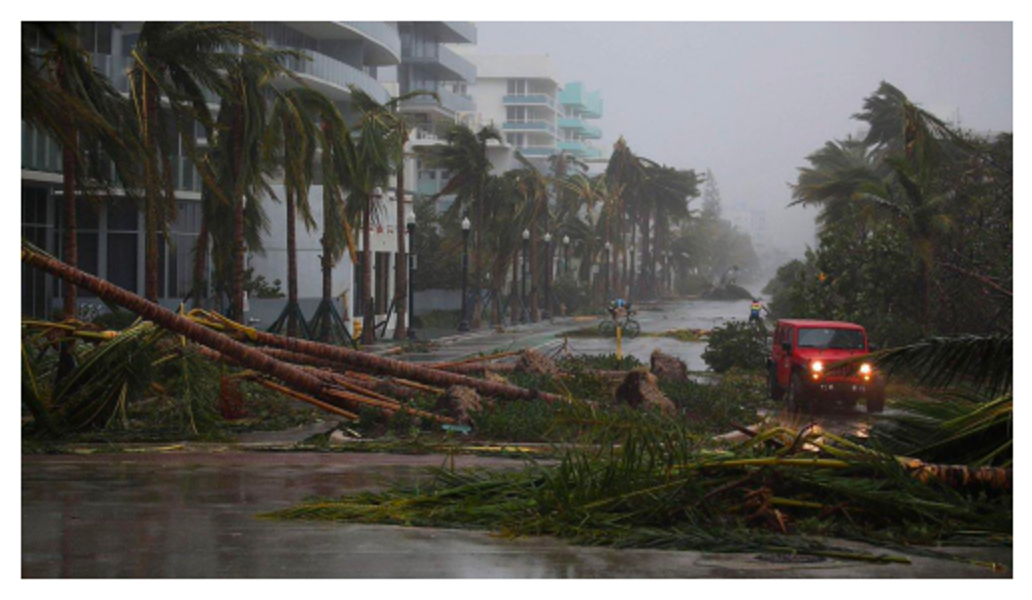}} & 
        Harvey, Irma could ding U.S. economy for combined \$290 billion \#finance
        & 
        
        The image shows a street filled with debris, including fallen trees and branches, and a street sign. There is also a truck and a person riding a bicycle in the scene. The street appears to be damaged due to a storm, possibly a hurricane, as suggested by the caption. The storm has caused significant damage to the area, including fallen trees and branches, which could potentially impact the local economy. The combined cost of the damage could be around \$290 billion, as indicated by the tweet. The image captures the aftermath of the storm, highlighting the need for recovery efforts and the potential economic consequences of such natural disasters.
        & mild damage \\
        \bottomrule
    \end{tabular}}
    \caption{Sample images and their corresponding tweet text as well as LLaVA-generated text for each of the tasks.}
    \label{fig:data_samples}
\end{figure*}

Fig.~\ref{fig:data_samples} provides sample examples of LLaVA-generated descriptions across three tasks: informativeness, humanitarian category, and damage severity. The model produces context-aware, detailed outputs by grounding visual features in tweet text, aligning well with ground-truth labels.

\begin{figure*}
  \centering
  \includegraphics[width=\linewidth]{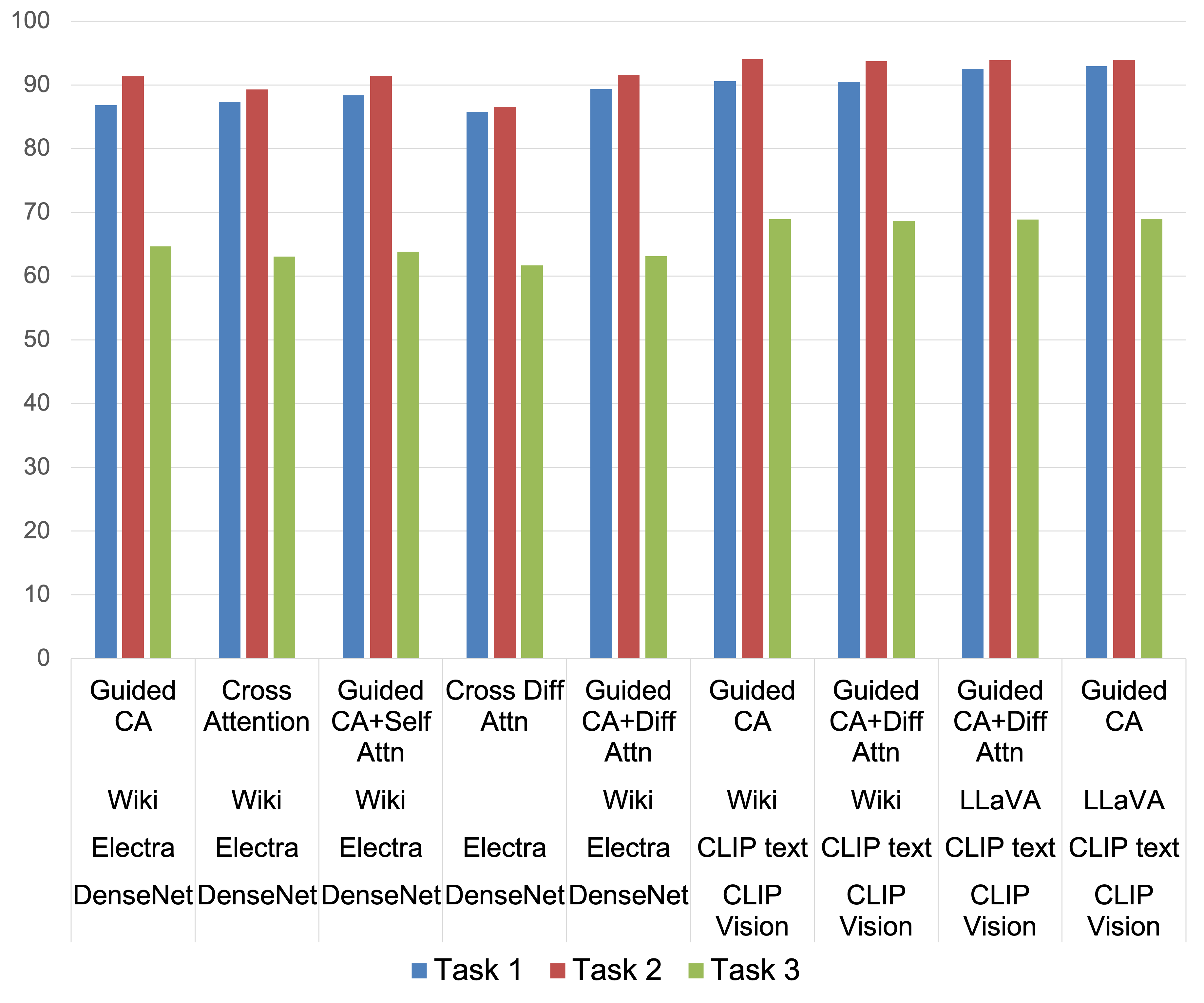}
  \hfill
  \caption{Accuracy across CrisisMMD tasks for vision and text fusion variants.}
  \label{fig:result_graph}
\end{figure*}

Fig.~\ref{fig:result_graph} compares model accuracy across three CrisisMMD tasks using various vision–text fusion strategies and backbone combinations. Guided cross-attention with CLIP consistently outperforms other variants, showing improved generalization across all tasks.

\end{document}